\documentclass[sigconf]{acmart}

\AtBeginDocument{%
  }

\copyrightyear{2024} 
\acmYear{2024} 
\setcopyright{acmlicensed}\acmConference[MMASIA '24]{ACM Multimedia Asia}{December 3--6, 2024}{Auckland, New Zealand}
\acmBooktitle{ACM Multimedia Asia (MMASIA '24), December 3--6, 2024, Auckland, New Zealand}
\acmDOI{10.1145/3696409.3700241}
\acmISBN{979-8-4007-1273-9/24/12}

\begin{document}

\title{GGAvatar: Reconstructing Garment-Separated 3D Gaussian Splatting Avatars from Monocular Video}

\author{Jingxuan Chen}
\email{jxc1424@student.bham.ac.uk}
\orcid{0009-0008-7078-509X}
\affiliation{%
  \institution{Jinan University-University of Birmingham Joint Institute}
  \city{Guangzhou}
  \state{Guangdong}
  \country{China}
}

\begin{abstract}
    Avatar modelling has broad applications in human animation and virtual try-ons. Recent advancements in this field have focused on high-quality and comprehensive human reconstruction but often overlook the separation of clothing from the body. To bridge this gap, this paper introduces \textbf{GGAvatar} (Garment-separated 3D Gaussian Splatting Avatar), which relies on monocular videos. Through advanced parameterized templates and unique phased training, this model effectively achieves decoupled, editable, and realistic reconstruction of clothed humans. Comparative evaluations with other costly models confirm GGAvatar's superior quality and efficiency in modelling both clothed humans and separable garments. The paper also showcases applications in clothing editing, as illustrated in Figure \ref{head}, highlighting the model's benefits and the advantages of effective disentanglement. The code is available at https://github.com/J-X-Chen/GGAvatar/.
\end{abstract}



\begin{teaserfigure}
  %\vspace{-0.3cm}
  \includegraphics[width=\textwidth]{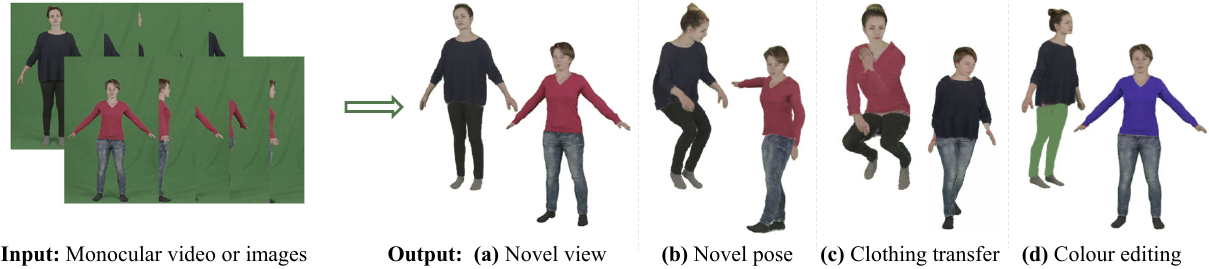}
  %\vspace{-0.2cm}
  \caption{This paper presents a novel editable avatar model with garments initialization and 3D Gaussian Splatting to achieve decouplable reconstruction of clothed humans. The model supports various applications, including standard novel view or pose synthesis tasks ((a), (b), (c)), (d)) and garment editing manipulations ((c)), (d)).}
  \label{head}
\end{teaserfigure}


\maketitle

\section{Introduction}
Reconstructing realistic clothed digital humans and their garments is a significant task in computer graphics and computer vision. This type of work aims to synthesize high-resolution clothed human body images from an unprecedented view or generate human imagery in a novel pose. Previous research has delved into explicit modelling methods under costly capture systems to obtain suboptimal reconstruction outcomes \cite{xianshi1, xianshi2}.  Recent advancements have shifted towards direct construction from single RGB images or monocular videos, utilizing models with implicit representation such as Neural Radiance Field (NeRF)\cite{nerf} to capture fine textures on the surface. However, these models \cite{neuralbody, humannerf, scarf, animnerf} require dozens of training hours. Consequently, current studies\cite{instantavatar,instant_nvr,3dgsavatar, gart, gaussianavatar, hugs,gauhuman} are increasingly focused on enhancing rendering speed and modelling efficiency by turning neural rendering techniques into Instant-NGP\cite{instantngp} or 3D Gaussian Splatting (3DGS)\cite{3dgs}. Nevertheless, the lack of disentanglement functions in these existing avatar models may constantly limit their applicability in real-world scenarios. This paper argues that an ideal avatar model should not only produce high-quality, rapid, and thorough reconstruction results, but also possess the decoupling capability necessary for applications such as virtual try-ons.

Unfortunately, creating a perfect editable and drivable avatar is a demanding task that presents several challenges. Firstly, to effectively disentangle the body and garments, integrity and anti-interference properties must be maintained between distinct components. Specific estimations are required for the unsupervised areas where the human body is obstructed. Secondly, a precise transformation between canonical space and various pose spaces must be established to locate the partitioned point cloud at the target position. Lastly, it is essential to capture diverse and intricate clothing details, including textures, and to achieve high-quality reconstructions from sparse monocular inputs, particularly for loose-fitting attire. However, works such as \cite{isp,delta,smplicit,bcnet,dig,gala,clothcap} are limited to recovering geometry without providing corresponding appearance information.

In response to these challenges, this paper proposes a novel framework, GGAvatar, designed to construct realistic avatars from monocular videos while effectively and completely separating the garments. Specifically, this paper builds and fits garment templates alongside the corresponding body template to achieve a preliminary state of separation and interference resistance, resulting in partitioned point sets. Phased trainable modules (isolation and joint training) reasonably prevent the intersection of point sets during the training process. Subsequently, the target Gaussian positions are ensured by constructing deformation fields based on a concentric skeleton. Simultaneously, high-quality rendering is accomplished using 3DGS. Notably, GGAvatar enables thorough separation of clothed humans in novel view synthesis tasks from monocular inputs—potentially a first in this field, to my knowledge.

The paper evaluates the GGAvatar model by comparing it with baseline approaches and other works on the People Snapshot Dataset \cite{peoplesnapshot} or the ZJU Mocap Dataset \cite{neuralbody}. The results indicate that GGAvatar demonstrates a high level of reconstruction quality for clothed humans, comparable to that of other 3DGS-based models. Notably, the proposed model outperforms nearly every traditional NeRF-based model while exhibiting significantly faster training speeds—approximately hundreds of times faster than the NeRF counterparts. Furthermore, ablation studies are conducted to validate the effectiveness of each component. To highlight the superiority and practical utility of GGAvatar, this paper compares it with existing non-fully decoupled models on clothing transfer.

The contributions are summarized as follows:
\begin{itemize}
\item This paper proposes the GGAvatar model, based on phased training methods, to achieve high-quality and efficient construction for various viewpoints or pose synthesis tasks of clothed humans.

\item  The method of constructing parameterized templates for garments is introduced to solve the challenge of complex clothes modelling.

\item  The GGAvatar enables a thorough separation between different garments, allowing applications such as colour editing and clothing transfer.
\end{itemize}

\section{Related Work}
\noindent{\normalfont{\textbf{Geometry Reconstruction.}}
The Skinned Multi-Person Linear (SMPL) model\cite{smpl} is a widely used parametric model for human body shape and pose estimation. The extended versions (such as SMPL-H, SMPL-X\cite{smplx}, STAR\cite{star}) can characterize the human body as a deformable mesh in a low-dimensional linear space with learned parameters, allowing for accurate capture of diverse body shapes and poses. Similarly, clothing geometry can be obtained from video sequences using methods\cite{clothcap,clothparam2,gala,dig,drapenet,bcnet,bmvc}. The SMPLicit\cite{smplicit} and Neural-ABC \cite{neuralabc} can effectively edit the modelled geometry of fully clothed humans with a latent vector. Additionally, the Implicit Sewing Patterns (ISP)\cite{isp} method enables efficient 3D reconstruction of multi-layered garments from a single image. In this approach, each garment mesh is reconstructed by sewing together two individual 2D panels associated with the Signed Distance Function (SDF) value of the 3D surface, alongside the stitching positions generated by neural networks. However, these models focus solely on geometric representation and ignore texture capture.

\hspace*{\fill}

\noindent{\normalfont{\textbf{Dynamic Character Reconstruction.}}}
HumanNeRF\cite{humannerf} and Neural Body\cite{neuralbody} leverage geometric priors to create implicit neural representations of dynamic humans, synthesizing realistic body details. InstantAvatar\cite{instantavatar} combines Instant-NGP\cite{instantngp} (a method that relies on CUDA programming and hash functions) with Fast-SNARF\cite{fastsnarf} (a spatial transformation algorithm), enabling rapid rendering. For the 3DGS-based techniques\cite{3dgs} (which map Gaussian point clouds directly instead of accumulating the colour values of the pixel blocks) the Gaussian Articulated Template Model (GART)\cite{gart},  Human Gaussian Splats (HUGS)\cite{hugs}, and GaussianAvatar\cite{gaussianavatar} enhance rendering speed and performance in the reconstruction of animations by managing loose garments with latent bones, constructing avatars and scenes, or performing pose pre-processing, respectively. Although these methods successfully reconstruct avatars, they all overlook the importance of disentanglement.

\hspace*{\fill}

\noindent{\normalfont{\textbf{Decouplable Human Reconstruction.}}} The existing separable models require training on a large set of 3D clothed human inputs obtained from multi-camera systems \cite{multivideo,multiview}. For instance, PhysAvatar \cite{physavatar} and LayGA \cite{layga} rely on multi-view video data to achieve excellent reconstruction through reverse and layered rendering frameworks. In contrast, the Segmented Clothed Avatar Radiance Field (SCARF) \cite{scarf}, an innovative hybrid model that merges SMPL-X \cite{smplx} with NeRF technology, allows for the reconstruction of clothed human avatars and their garments directly from monocular videos. On the basis of SCARF, Disentangled Avatars (DELTA) \cite{delta} builds an additional hair component while maintaining the same garment modelling method. However, these approaches have drawbacks, including extensive training time requirements and limited capacity for separation between garments.

\section{Parametric Representation}

\begin{figure*}[t]
  \includegraphics[width=\textwidth]{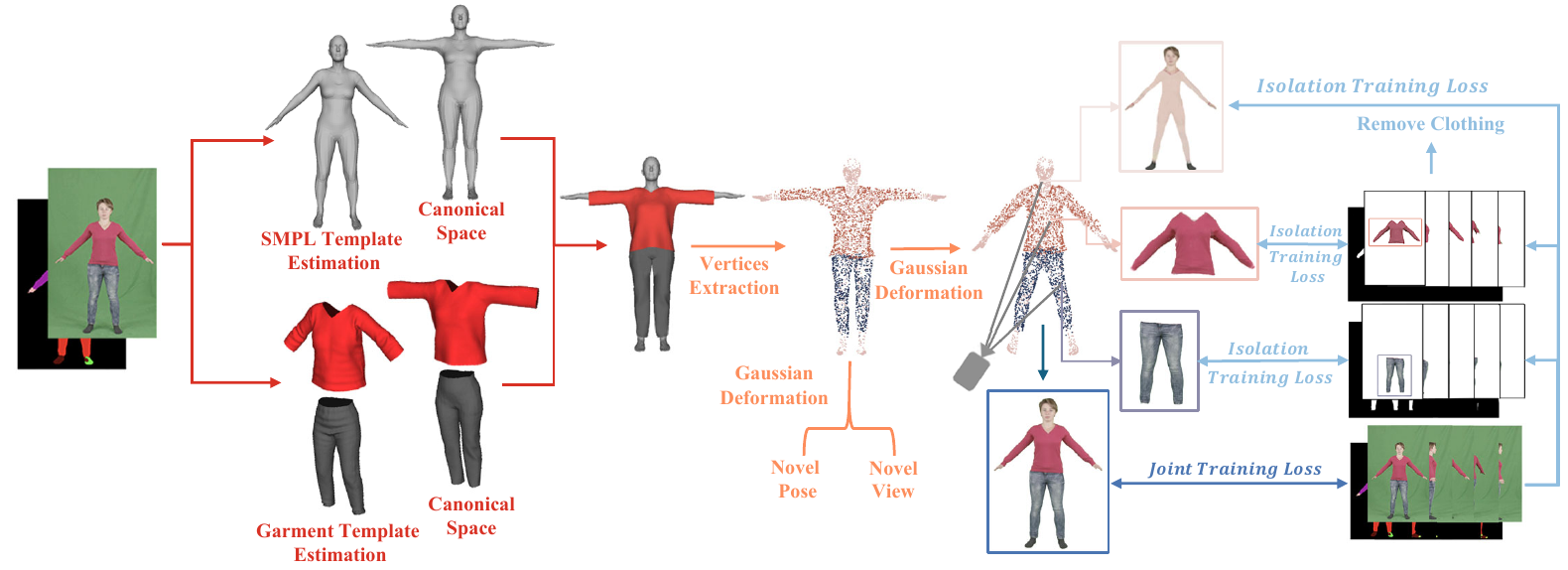}
  \caption{The Framework of GGAvatar. The model accepts monocular RGB images or videos along with their masks as input, producing an editable clothed human avatar that can be viewed from any perspective for any pose. The red line symbolizes the garment template modelling procedure, the orange line represents the deformation process, and the blue line indicates two types of rendering optimization.}  
  \vspace{-0.1cm}
\label{pipeline}
\end{figure*}

\noindent{\normalfont{\textbf{Body Representation.}}}
The SMPL\cite{smpl} model $\mathbf{T}^{(b)}({\beta},{\theta})$ with shape $\beta \in \mathbb{R}^{10}$ and pose $\theta \in \mathbb{R}^{(n_k+1) \times 3}$ is utilized as a low-cost parametric expression of the human body, where $n_{k}$ ($n_{k}=23$) is the number of the joints. Following \cite{delta, scarf}, this paper adds offsets  $\mathrm{O} \in \mathbb{R}^{|\mathcal{V}^{(b)}| \times 3} $ to all the $|\mathcal{V}^{(b)}|$ vertices to capture localized geometric details. The human template ($\mathcal{V}^{(b)}, Face^{(b)}$) can be totally represented as follows.

\begin{equation}
\begin{aligned}
    \mathbf{M}^{(b)}({\beta}, {\theta})&=W\left(\mathbf{T}_{can}^{(b)}({\beta},{\theta},O), J({\beta}), {\theta}, \mathcal{W}^{(b)} \right),\\
    \mathbf{T}_{can}^{(b)}({\beta},{\theta})&=\overline{\mathbf{T}}^{(b)}+B_{s}(\beta)+B_{p}(\theta) +O.
\end{aligned}
\end{equation}
Here, the $J(\beta) \in \mathbb{R}^{|\mathcal{V}^{(b)}| \times 3}$ can establish a correspondence between the skeletal joints and the mesh vertices, $\overline{\mathbf{T}}^{(b)} \in \mathbb{R}^{|\mathcal{V}^{(b)}| \times 3}$ is a template in rest pose provided by \cite{smpl} and the blend skinning weights  $\mathcal{W} \in \mathbb{R}^{n_k \times |\mathcal{V}^{(b)}|}$ is used for deforming according to linear blend skinning (LBS).

\hspace*{\fill}

\noindent{\normalfont{\textbf{Garment Representation.}}}
Similar to the existing methodologie\cite{scarf} that represents various types of clothing in wild videos, the $c$-th garments (where $c \in \{1,2,\dots,n_c\}$) can be attached to the body template using the human pose $\theta$, human shape $\beta$, joint correspondence $J({\beta})$, and deformation methods, but with distinct skin weights. The equation governing this process is as follows.
\begin{equation}
    \mathbf{M}^{(c)}(\beta,\theta)=W\left(\mathbf{T}_{can}^{(c)}(\beta,\theta), J({\beta}), {\theta}, \mathcal{W}^{(c)} \right).
\end{equation}
It can be observed that alignment can be achieved due to the consistent SMPL template parameters, ensuring the same topological deformation for garments and their underlying body. The generation of the template $\mathbf{T}^{(c)}_{can}(\beta,\theta) \in \mathbb{R}^{|\mathcal{V}^{(c)}| \times 3}$ will be elaborated upon in Section \ref{garment initialization}.

\section{Method}
This module offers a comprehensive overview of the GGAvatar model, illustrated in Figure \ref{pipeline}. The entire process can be systematically divided into three sections: garment initialization, deformation field processing and rendering.

\subsection{Garment Templates Estimation}
\label{garment initialization}
Currently, some avatar models\cite{3dgsavatar,gaussianavatar} adopt a points offset strategy to depict the position of the outermost single layer of clothing. However, reconstructing clothing based solely on a single layer is insufficient for effective garment isolation, as a clothed human typically involves multiple layers of geometry (at least one layer of skin and one layer of garment). Therefore, additional clothing templates should be employed as an initialization stage for GGAvatar, tackling the modelling challenges posed by garment irregularities.

The garment template estimation begins with accurately estimating the human pose using FrankMocap\cite{frankmocap} to determine the correct parameters $\beta$ and $\theta$. Due to variations in SMPL estimation across different views, the front view of the human body must be selected and aligned with the corresponding segmentation from Self Correction for Human Parsing (SCHP) method\cite{schp}, serving as the input for the Implicit Surface Prediction (ISP) model\cite{isp}. After applying two types of Multi-layer Perceptrons (MLPs), the front and back clothing components, with SDF values and 3D shapes, are merged according to stitch-up order numbers to create a clothing mesh as $\mathbf{T}^{(c)}(\beta,\theta)$.

It is essential to establish correct spatial alignment between all garments and the human body. All the garment templates are stored within a unified canonical space, defined as $\mathbf{T}^{(c)}_{can}(\beta,\theta)$, which is consistent with the human body's SMPL model. After reshaping and building, vertices are extracted from these standard meshes to serve as the initial positions for garment Gaussian representation.

\subsection{Gaussian Representation and Deformation}
Drawing inspiration from 3D Gaussian Mixture Models\cite{gart}, both garments and the human body reconstruction results should be represented as Gaussians. The Gaussian vertex set $\mathcal{V}$ can be $\mathcal{V}^{(c)}, \mathcal{V}^{(b)}$ or a combination of these. The $i$-th Gaussian component in Gaussian set with vertices $\mathcal{V}$ can be defined by a 3D mean $\mu^{(i)}$, a 3D rotation $R^{(i)}$ for orienting, anisotropic scaling factors $s^{(i)}$ for size adjusting, an opacity factor $\eta^{(i)}$ and a colour radiance function expanded by spherical harmonics \cite{3dgs,plenoxels}. Additionally, the learnable skinning weights $\widetilde{\mathcal{W}_k}$ for the $k$-th joint, with adjustment $\Delta w_j$, can be written as:
\begin{equation}
    \label{weights}
    \widetilde{\mathcal{W}}_k (\mu^{(i)})=\mathcal{W}_k (\mu^{(i)})+\Delta w_j,\quad i \in \mathcal{V},\  k \in \{1,2,\dots,n_k\},
\end{equation}
where $\widetilde{\mathcal{W}}_k (\mu^{(i)}) \in \mathbb{R}$ is in the canonical space. Utilizing the learnable skinning weights and target bone transformations $\mathcal{B}=\{B_{1}, B_{2}, \ldots, $ $B_{n_{k}}\}$ for rigid transformation\cite{gart,instantavatar,smpl}, the surface Gaussians in the canonical space can be deformed to $\mu'$ in any observation space via the LBS shown as follows.
\begin{equation}
    \mathcal{T}^{(i)} = \sum_{k=1}^{n_k} \widetilde{\mathcal{W}}_{k}(\mu^{(i)}) B_{k}, \quad \mu' = \widetilde{\mathcal{T}}^{(i)}_{(1:3,1:3)} \mu^{(i)} + \widetilde{\mathcal{T}}^{(i)}_{(4,1:3)}.
\end{equation}
To simplify the representation, a weight matrix $\mathcal{T}^{(i)}$ is introduced, which is structured such that the upper left submatrix ($\mathbb{R}^{3\times3}$) corresponds to rotation, while the right submatrix ($\mathbb{R}^{1\times3}$) represents translation. This weight matrix applies to other Gaussian attributes representing Gaussian sets within any other observation space.
\begin{equation}
\begin{aligned}
    &\text{In the canonical space: } \\
    & \quad \mathcal{G}_0(\theta):=\{\mu^{(i)},R^{(i)}, s^{(i)}, \eta^{(i)}, f^{(i)}\}_{i=1}^{|\mathcal{V}|}, \\
    &\text{In the observation space: } \\
    & \quad \mathcal{G}(\theta):=\{\mu'^{(i)},R'^{(i)}, s^{(i)}, \eta^{(i)}, f^{(i)}\}_{i=1}^{|\mathcal{V}|} \\
    & \quad \quad \quad \ =\{\widetilde{\mathcal{T}}^{(i)}_{(1:3,1:3)}\mu^{(i)} + \widetilde{\mathcal{T}}^{(i)}_{(4,1:3)}, \widetilde{\mathcal{T}}^{(i)}_{(1:3,1:3)}R^{(i)}, s^{(i)}, \eta^{(i)}, f^{(i)}\}_{i=1}^{|\mathcal{V}|}.
\end{aligned}
\end{equation}

By manipulating the implicit skeleton, the Gaussian sets from all garments and the human body are assigned new coordinates in the image pose space while maintaining the alignment relationship and other Gaussian attributes. This is attributed to a mutually independent deformation with the same skeleton. Another task of the deformation field is to ensure that a bordered state is upheld across Gaussian sets throughout the training process, leading to a clear differentiation among various entities in the final output.

\subsection{Rendering with Splatting.}
To preserve the correct shape and orientation of the Gaussian, the covariance is defined as $\Sigma^{(i)} = R S S^{T} R^{T}$, where $S$ is the scaling matrix and $R$ is the rotation matrix, effectively characterizing the Gaussian ellipsoid. In the context of a camera with an extrinsic $E$ and an intrinsic matrix $K$, 3DGS\cite{3dgs} achieves notable quality and speed enhancements by employing the mapping operation $\pi(x; E, K)$, in contrast to the ray-by-ray calculations used in NeRF \cite{nerf}. Let $\mathcal{C}^{(i)}_{\theta}\left(x, d\right)$ denote the colour at querying position $x$ from the $i$-th Gaussian in the observation space. Following most volume rendering techniques and drawing inspiration from NeRF and Radon transformation theory, the rendered image can be generated from 2D observations as follows.
\begin{equation}
\begin{aligned}
    \mathcal{I}(\mathcal{G}(\theta), d) &=\sum_{i\in N} \mathcal{C}^{(i)}_{\theta}\left(x, d\right)\alpha^{(i)}\prod_{j=1}^{i-1}\left(1-\alpha^{(j)}\right),\\
%    \alpha^{(i)} &=G_{2D}\left(u \mid \eta^{(i)}, \mu_{2D}^{(i)}, \Sigma_{2 D}^{(i)}\right),
    \mathcal{C}^{(i)}_{\theta}\left(x, d\right)&=\operatorname{sph}\left({R'^{(i)}}^{T} d, f^{(i)}\right).
\end{aligned}
\end{equation}
The value of $\alpha^{(i)}$, as the contribution of the final colour, depends on the position, opacity, and covariance matrix of the current $i$-th Gaussian component. It is computed using a 2D Gaussian ($\mu_{2d}^{(i)}=\pi(\mu^{(i)}; E, K), \ \Sigma_{2d}^{(i)}=J E \Sigma^{(i)} E^{T} J^{T},\ J \in \mathbb{R}^{2\times3}$). The opacity multiplication for each point follows the procedures discussed in previous works \cite{ewa, gart, 3dgs}.

Due to varying forms of supervision during different training periods, the notation $N$ can be specified as $N^{(b)}$, $N^{(c)}$ (distinct from $|\mathcal{V}^{(b)}|$, $|\mathcal{V}^{(c)}|$) or together, representing the number of overlapping Gaussians in a specified depth order in the $d$-direction across different training phases. All processes are differentiable, enabling back-propagation and optimization. 

\subsection{Training Losses}
\label{loss}
This paper augments the baseline optimization \cite{3dgs,gart} with additional regularization terms to ensure a smooth effect. In the initial isolation stage, a densify-and-prune strategy, as outlined in 3DGS, is employed and every reconstruction loss for garments and body parts under the SCHP \cite{schp} segmentation is summed up. In the subsequent joint training phase, this paper optimizes the Gaussians without adding or removing any components. The primary reconstruction loss\cite{plenoxels,3dgs,gart} between ground truth $\mathcal{I}(\theta)$ and rendered image $\mathcal{\hat{I}}(\theta)$ can be expressed as follows for each estimated pose $\theta$. 
\begin{equation}
    \mathcal{L}_{recon} = \mathcal{L}_{1}(\mathcal{\hat{I}}(\theta),\mathcal{I}(\theta)).
\end{equation}

\noindent{\normalfont{\textbf{Stochastic Structural Similarity Loss.}}}
Considering distant or global information, the model replaces the Structural Similarity Index measure (SSIM) loss\cite{ssim} with the Stochastic Structural SIMilarity (S3IM) loss\cite{s3im}. The rendered image is randomly cropped into the patches $\mathcal{\hat{I}}^{(n)}_{\mathrm{patch}}(\theta)$ with the same operation applied to the corresponding ground-truth image patches $\mathcal{I}^{(n)}_{\mathrm{patch}}(\theta)$ to calculate SSIM using a $K \times K$ kernel and a stride of $s$.
\begin{equation}
    \mathcal{L}_{\mathrm{S3IM}} = \frac{1}{N_r} \sum_{n=1}^{N_r} \mathcal{L}_{\mathrm{SSIM}}\left(\mathcal{\hat{I}}^{(n)}_{\mathrm{patch}}(\theta), \mathcal{I}^{(n)}_{\mathrm{patch}}(\theta)\right),
\end{equation}
where $N_r$ is the number of repetitions.

\hspace*{\fill}

\noindent{\normalfont{\textbf{Deformation Regularization.}}}
Inspired by \cite{dynamic3dgs} and \cite{geometric}, this paper leverages the "as-isometric-as-possible" constraint to preserve a similar distance and shape after deformation, thereby reducing the occurrence of artefacts.
\begin{equation}
\begin{aligned}
    \mathcal{L}_{iso} = \sum_{i=1}^{|\mathcal{V}|} \sum_{j \in \mathcal{N}(i)}
    &\left(\lambda_{\mu} \left|d\left(\mu^{(i)}, \mu^{(j)}\right) - d\left(\mu_{o}^{(i)}, \mu_{o}^{(j)}\right)\right| \right.\\
    &+\left. \lambda_{\Sigma} \left|d\left(\boldsymbol{\Sigma}^{(i)}, \boldsymbol{\Sigma}^{(j)}\right) - d\left(\boldsymbol{\Sigma}_{o}^{(i)}, \boldsymbol{\Sigma}_{o}^{(j)}\right)\right|\right).
\end{aligned}
\end{equation}
In this equation, the distance function $d$ corresponds to the $L_2$ norm. Subscripts of zero denote attribute values under the canonical pose. 

\begin{table*}[t]
\footnotesize
\centering 
  \caption{A quantitative comparison between GGAvatar and other existing avatar models (Neural Body\cite{neuralbody}, InstantAvatar\cite{instantavatar}, GaussianAvatar\cite{gaussianavatar}, GART\cite{gart}, SCARF\cite{scarf}), where the symbol "\dag" indicates the model possesses decoupling capabilities. Higher PSNR and SSIM, and lower LPIPS are desirable. The best score is highlighted in \textbf{bold}, while the second-best is \underline{underlined}.}
  \label{tab: avatar}
  \setlength\tabcolsep{3pt}
  \resizebox{\linewidth}{!}{
  \begin{tabular}{c|c|c|c|c}
    \toprule
    \small Methods (time) & \small male-3-casual & \small male-4-casual & \small female-3-casual & \small female-4-casual\\ 
    \midrule
    & {\footnotesize PSNR SSIM LPIPS}&{\footnotesize PSNR SSIM LPIPS}&{\footnotesize PSNR SSIM LPIPS}&{\footnotesize PSNR SSIM LPIPS}\\
    \small Neural Body($>$0.5d)& {\footnotesize24.94 0.9428 0.0326} & {\footnotesize 24.71 0.9469 0.0423} & {\footnotesize 23.87 0.9504 0.0346} & {\footnotesize 24.37 0.9451 0.0382} \\
    \small InstantAvatar($\sim$5m) & {\footnotesize29.53 0.9716 \underline{0.0155}} & {\footnotesize27.67 0.9626 0.0307} &{\footnotesize27.66 0.9709 \textbf{0.0210}} & {\footnotesize29.11 0.9683 \underline{0.0167}}\\
    \hspace{-0.3cm} \small GaussianAvatar($\sim$0.5h)& {\footnotesize\underline{30.98} \underline{0.9790} \textbf{0.0145}} & {\footnotesize28.78 \textbf{0.9755} \textbf{0.0228}} & {\footnotesize\underline{29.55} \underline{0.9762} \underline{0.0225}} &  {\footnotesize\textbf{30.84} \textbf{0.9771} \textbf{0.0140}} \\
    \small GART($\sim$2m)& {\footnotesize30.73 0.9769 0.0360} & {\footnotesize27.58 0.9679 0.0602} &{\footnotesize26.60 0.9659 0.0466} & {\footnotesize29.23 0.9721 0.0378} \\
    \midrule
    \small SCARF($>$2d)\dag & {\footnotesize 30.59 0.9770 0.0247} & {\footnotesize \underline{28.99} 0.9701 \underline{0.0257}} & {\footnotesize \textbf{30.14} \textbf{0.9764} 0.0282} & {\footnotesize 29.96 0.9718 0.0267} \\
    \small SCARF($\sim$20m)\dag & {\footnotesize 18.20 0.8521 0.3578} & {\footnotesize 18.41 0.8860 0.3618} & {\footnotesize 19.90 0.9120 0.2494} & {\footnotesize 19.76 0.8920 0.2274} \\
    \midrule
    \small GGAvatar($\sim$20m)\dag & {\footnotesize \textbf{31.01} \textbf{0.9812} 0.0349} & {\footnotesize \textbf{29.40} \underline{0.9728} 0.0509} & {\footnotesize 27.96 0.9753 0.0452} &{\footnotesize \underline{30.29} \underline{0.9750} 0.0343}\\
    \bottomrule
  \end{tabular}
  }
\end{table*}

\hspace*{\fill}
The complete loss function comprises several components: an RGB loss $\mathcal{L}_{recon}$, a mask loss $\mathcal{L}_{mask}$ for both garments and body,  $\mathcal{L}_{S3IM}$, $\mathcal{L}_{iso}$, a Gaussians similarity regularization $\mathcal{L}_{G\_reg}$ and a collision loss $\mathcal{L}_{col}$ for the joint training. Overall, the loss function of these two training stages is manifested as
\begin{equation}
\begin{aligned}
    &\mathcal{L}_{isolation}=\sum_{c}^{n_c}\mathcal{L}_{recon}^{(c)} + \lambda_{2} \times \mathcal{L}_{mask}^{(c)} + \lambda_{3} \times \mathcal{L}_{S3IM}^{(c)} + \lambda_{4} \times \mathcal{L}_{G\_reg}^{(c)},\\
    &\mathcal{L}_{joint}=\mathcal{L}_{recon} + \lambda_{2} \times \mathcal{L}_{mask} + \lambda_{3} \times \mathcal{L}_{S3IM} \\
    &\quad \quad \quad \quad \quad \quad \ \ \ + \lambda_{4} \times \mathcal{L}_{G\_reg} + \lambda_{5} \times \mathcal{L}_{iso} + \lambda_{6} \times \mathcal{L}_{col}.
\end{aligned}
\end{equation}
For more details on the experimental process, please refer to the supplementary materials.

\section{Experiments}
\begin{figure}[t]
  \includegraphics[width=0.5\textwidth]{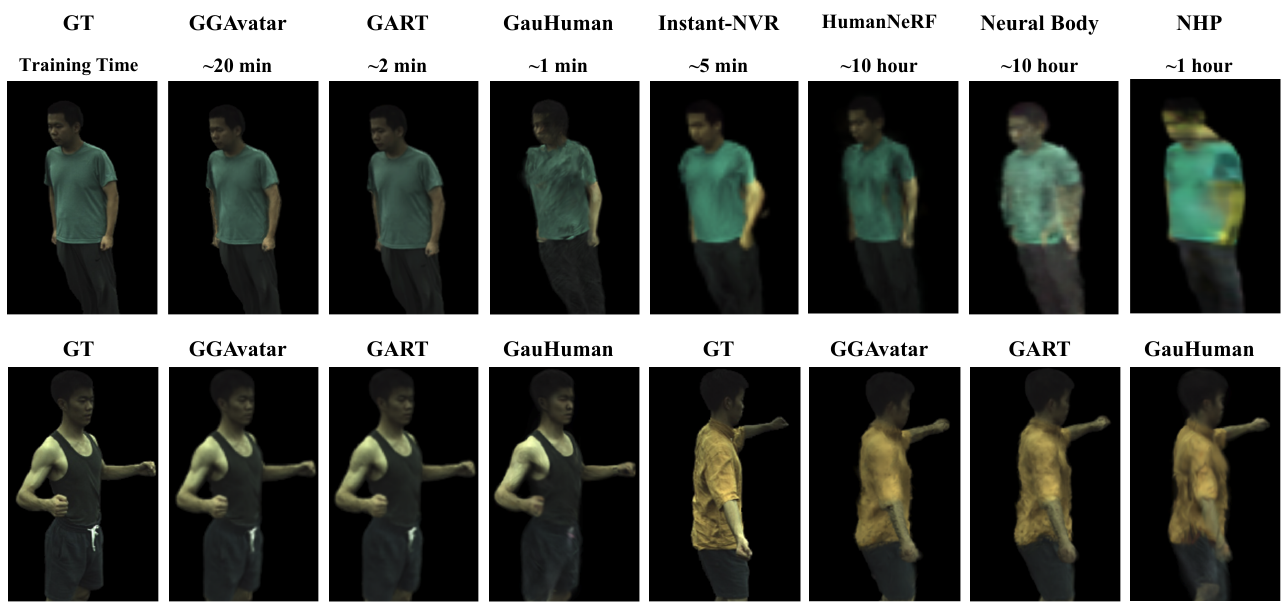}
  \caption{Results of novel view synthesis on the ZJU-MoCap.}
\vspace{-0.3cm}
\label{zju_compare}
\end{figure}
This paper performs a comprehensive evaluation and application demonstration of the proposed model on monocular datasets. The effectiveness of GGAvatar will be assessed by comparing it with existing Nerf-based and 3DGS-based methods under the same SMPL pose, along with an ablation test.

\subsection{Dataset}
The People Snapshot dataset \cite{peoplesnapshot} consists of videos of individuals rotating in front of a stationary camera, while the ZJU-MoCap dataset \cite{neuralbody} provides detailed motion sequences of various human activities, capturing a range of poses, expressions, and clothing styles. The People Snapshot dataset is split into 80\% for training, 10\% for validation, and 10\% for testing. In the case of the ZJU-MoCap dataset, the experiments use images from 'camera 1' as input and employ the other cameras for evaluation.

\subsection{Comparison with SoTA Methods}
\noindent{\normalfont{\textbf{Whole Human Synthesis Quality.}}
To compare the proposed model with existing non-couplable methods and assess its adaptability to novel perspectives or poses, numerous avatar methodologies were thoroughly trained for the recommended number of epochs as outlined in the reference papers. Table \ref{tab: avatar} showcases the improvement in certain image metrics (PSNR, SSIM\cite{ssim}, and LPIPS\cite{lpips}) for GGAvatar on the People Snapshot dataset, outperforming methods related to NeRF, such as Neural Body\cite{neuralbody}, InstantAvatar\cite{instantavatar}, and the baseline GART\cite{gart}. Furthermore, Figure \ref{zju_compare}, qualitatively analyzes the resemblance between the synthetic images and the ground truth on the ZJU-MoCap dataset. Notably, the clarity of the results produced by the proposed model is superior to that of competing methods. In summary, the reconstruction results can reach state-of-the-art performance in certain metrics across specific datasets, positioning  GGAvatar on par with other 3DGS-based methods.

\hspace*{\fill}

\begin{figure}[t]
\centering
\includegraphics[width=0.8\linewidth]{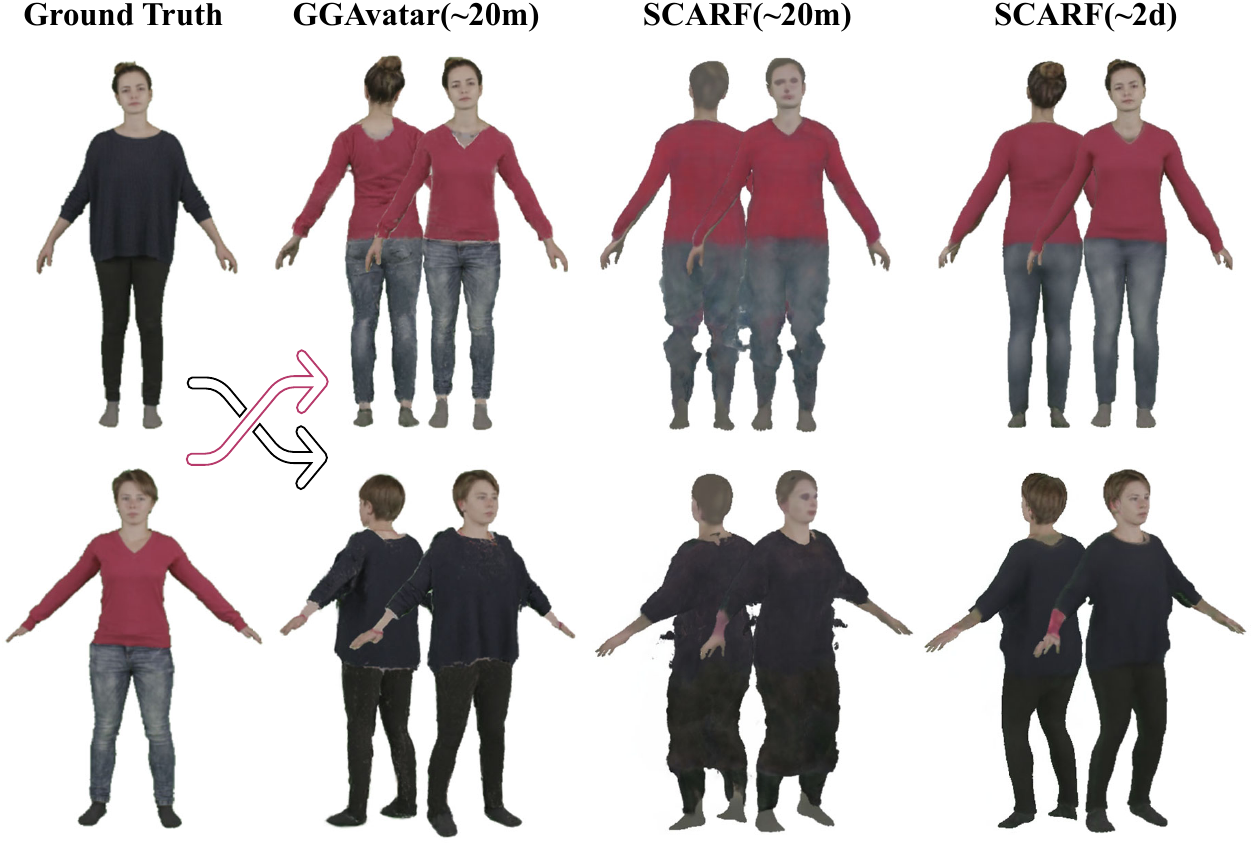}
%\vspace{0.1cm}
\caption{The results of holistic clothing transfer between GGAvatar and SCARF\cite{scarf} (under two different training times). The modelling garments are displayed by attaching them to another person on the People Snapshot dataset\cite{peoplesnapshot}.}
\vspace{-0.2cm}
\label{scarf_compare}
\end{figure}

\noindent{\normalfont{\textbf{Garment Reconstruction Quality.}}}
Considering the scarcity of models for reconstructing texture-based clothing from monocular videos, this paper highlights the quality of clothing reconstruction primarily through comparison with the SCARF\cite{scarf}, which is equivalent to the Delta\cite{delta} model. Figure \ref{scarf_compare} illustrates that clothing reconstruction surpasses both versions of SCARF (the 20-minute and 2-day iterations). Under the same time constraints, SCARF fails to achieve satisfactory clothing modelling within 20 minutes. In comparison to the full model, the results demonstrate rationality at the clothing junctions, while SCARF exhibits a noticeable colour gradient effect. Moreover, human skin processed by SCARF shows significant staining from clothing, which is much more pronounced than in the proposed approach.

\hspace*{\fill}

\noindent{\normalfont{\textbf{Reconstruction Speed.}}}
GGAvatar requires a mere 20-minute training duration on a single RTX 3080 Ti (laptop or server), which is significantly faster than NeRF-based methods. For instance, Neural Body\cite{neuralbody} demands 14 hours on $4\times$ RTX 2080 or a full training day on a 3080 Ti, while HumanNeRF\cite{humannerf} mandates approximately 3 days on the same hardware setup. For the decoupled models, SCARF\cite{scarf} entails around 2 days (40 hours on NVIDIA V100) for training, which is hundreds of times slower than GGAvatar. Additionally, it achieves remarkable rendering efficiency, delivering approximately 80 FPS for images sized at $540\times540$.

\begin{table}[t]
\centering
\caption{An Ablation of the loss function and joint training strategy of GGAvatar on female-4-causal.}
\begin{tabular}{c|cccc}
    \toprule
    \small Methods&\small PSNR$\uparrow$&\small SSIM$\uparrow$&\small LPIPS$\downarrow$&\small\\
    \midrule
    \small Full Model&30.29&0.9750&0.0343&\small\\
    \small w/o $\mathcal{L}_{\mathrm{S3IM}}$&29.81&0.9731&0.0307&\small\\
    \small w/o $\mathcal{L}_{iso}$&30.27&0.9749&0.0347&\small\\
    \small w/o Joint Training&28.29&0.9702&0.0361&\small\\
  \bottomrule
\end{tabular}
\label{tab: ablation}
\end{table}
\begin{figure}[]
\centering
\vspace{-0.1cm}
\includegraphics[width=0.7\linewidth]{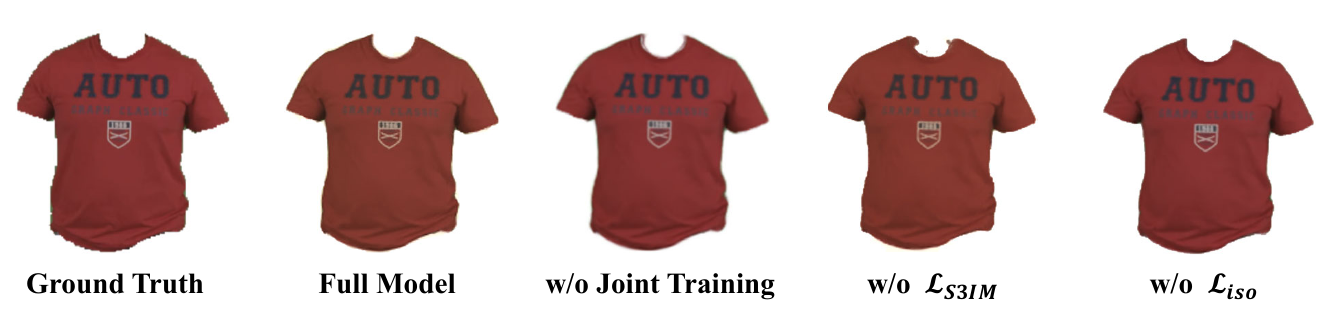}
%\vspace{0.1cm}
\caption{Ablation results for the clothes from male-3-causal are obtained. Due to consistent Gaussian rendering methods employed, the middle regions in the figure showcase similar reconstruction effects.}
\vspace{-0.5cm}
\label{fig: ablation}
\end{figure}

\hspace*{\fill} 

\subsection{Ablation Study}
Table \ref{tab: ablation} and Figure \ref{fig: ablation} summarize an ablation study conducted to assess the influence of training type and loss function. The differences are primarily reflected in the metrics (PSNR, SSIM, LPIPS) and the contours around the outputs.

\hspace*{\fill} 

\noindent{\normalfont{\textbf{Training strategy.}}}
The investigation of the multiple training steps, as illustrated in Table \ref{tab: ablation} and Figure \ref{fig: ablation}, indicates that the model lacking the final joint training exhibits performance deficiencies. This reduction may be attributed to the imperfect merging of clothing boundaries and the human body during the stitching process, resulting in artefacts at the connections.

\hspace*{\fill} 

\noindent{\normalfont{\textbf{Loss function.}}}
As depicted in Table \ref{tab: ablation} and Figure \ref{fig: ablation}, the omission of both $\mathcal{L}_{\mathrm{S3IM}}$ and $\mathcal{L}_{iso}$ leads to a decline in rendering quality. The regularization loss $\mathcal{L}_{\mathrm{S3IM}}$, as described in Section \ref{loss}, effectively reduce noise surrounding the 3D contours of the garments in Figure \ref{fig: ablation}.

\begin{figure}[t]
\vspace{-0.2cm}
    \centering
\includegraphics[width=0.7\linewidth]{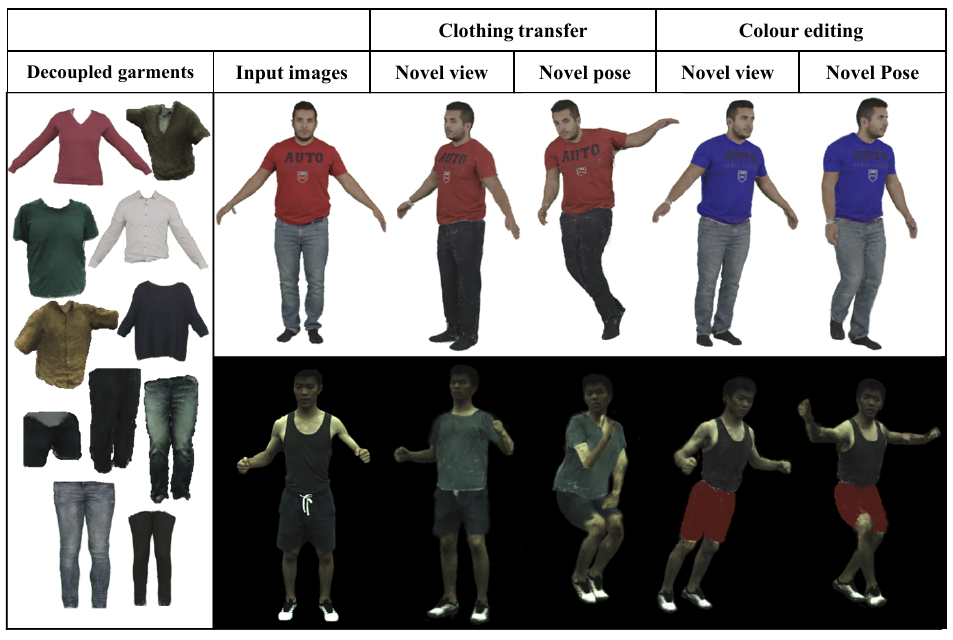}
\caption{The clothing transfer and the colour-changing application of GGAvatar.}
\vspace{-0.3cm}
\label{application}
\end{figure}

\subsection{Application}
On the application level, in addition to the basic function of synthesising new perspective images, GGAvatar offers three extra features described in this section.

\hspace*{\fill}

\noindent{\normalfont{\textbf{Animation.}}}
Similar to recent avatar approaches, GGAvatar offers fine-grained control over novel body poses and movements, such as eating, walking, and dancing. Figure \ref{application} shows that the model preserves reconstruction quality even when animations are driven according to predefined action sequences

\hspace*{\fill}

\noindent{\textbf{Clothing Transferring.}}
The model facilitates clothing transfer across avatars by combining Gaussians from diverse individuals. Unlike methods that only support holistic garment transfer, such as SCARF, this approach achieves inter-clothing separation, allowing the transfer of individual clothing items, such as shirts or pants, to another human, as demonstrated in the middle section of Figure \ref{application}. Furthermore, there is flexibility in selecting distinct garment components, as shown on the left side of the figure.

\hspace*{\fill}

\noindent{\normalfont{\textbf{Colour Editing.}}}
Regarding clothing colour modifications, the model accepts colour keywords or RGB values as input without necessitating manual conversion of spherical harmonics. For example, Figure \ref{application} illustrates the alteration of the pants colour from subject 377 (from the ZJU-MoCap dataset) to crimson (RGB $= [80,0,0]$) and swapping of the two colour channels of male-3-casual (from the People Snapshot dataset) to achieve either localized colour adjustment or a reversed colour effect.

\section{Conclusion}
This paper presents GGAvatar models to realise the separation of humans and garments while maintaining high-fidelity holistic avatar reconstruction. Unlike the existing decoupling models, GGAvatar improves reconstruction speed, adequacy of garment decoupling and overall quality through clothing initialisation, separation-based training and optimisation. Extensive experiments on the novel view or pose synthesis consistently demonstrate that the proposed model surpasses most implicitly represented clothed human reconstruction models in terms of both quality and efficiency. Additionally, this work proves the effectiveness of clothing editing. In real-world applications, the decoupled garments and novel views display features that make GGAvatar well-suited for virtual reality and virtual try-on scenarios.

\clearpage

\bibliographystyle{ACM-Reference-Format}
\bibliography{sample-base-cjx}


\end{document}